\documentclass[]{fairmeta}
\usepackage{amsmath}
\usepackage{amsthm}
\usepackage{amssymb}
\usepackage{mathtools}  % Enhancements to amsmath (e.g., \coloneqq)

\usepackage[most]{tcolorbox}
\usepackage{tabularx}
\usepackage{xcolor}
\usepackage{graphicx}
\usepackage{booktabs}
\usepackage{caption}

\usepackage{wrapfig}
\usepackage{graphicx}

\usepackage{xcolor}
\usepackage{mdframed}
\usepackage{tabularx, booktabs}

\definecolor{lightgray}{gray}{0.95}
\definecolor{lightblue}{RGB}{220,235,255}
\definecolor{midblue}{RGB}{60,90,160}
\definecolor{tealbar}{HTML}{56C4C0}
\definecolor{violetbar}{HTML}{A245EB}

\newmdenv[
  backgroundcolor=lightgray,
  linecolor=midblue,
  linewidth=1pt,
  skipabove=8pt,
  skipbelow=8pt,
  roundcorner=3pt,
  innertopmargin=6pt,
  innerbottommargin=6pt,
  innerleftmargin=8pt,
  innerrightmargin=8pt
]{graybox}

\newmdenv[
  backgroundcolor=lightblue,
  linecolor=midblue,
  linewidth=0.8pt,
  skipabove=6pt,
  skipbelow=6pt,
  roundcorner=2pt,
  innertopmargin=6pt,
  innerbottommargin=6pt,
  innerleftmargin=8pt,
  innerrightmargin=8pt,
]{bluebox}

\usepackage[ruled,vlined]{algorithm2e}

\title{Bi-Level Prompt Optimization for Multimodal LLM-as-a-Judge}

\author[1,2,*]{Bo Pan}
\author[1,\dagger]{Xuan Kan}
\author[1]{Kaitai Zhang}
\author[1]{Yan Yan}
\author[1]{Shunwen Tan}
\author[1]{Zihao He}
\author[1]{Zixin Ding}
\author[1]{Junjie Wu}
\author[2]{Liang Zhao}

\affiliation[1]{Meta AI}
\affiliation[2]{Emory University}

\contribution[*]{Work done at Meta}
\contribution[\dagger]{Corresponding Author}

\abstract{
Large language models (LLMs) have become widely adopted as automated judges for evaluating AI-generated content. Despite their success, aligning LLM-based evaluations with human judgments remains challenging. While supervised fine-tuning on human-labeled data can improve alignment, it is costly and inflexible, requiring new training for each task or dataset. Recent progress in auto prompt optimization (APO) offers a more efficient alternative by automatically improving the instructions that guide LLM judges. However, existing APO methods primarily target text-only evaluations and remain underexplored in multimodal settings. In this work, we study auto prompt optimization for multimodal LLM-as-a-judge, particularly for evaluating AI-generated images. We identify a key bottleneck: multimodal models can only process a limited number of visual examples due to context window constraints, which hinders effective trial-and-error prompt refinement. To overcome this, we propose BLPO, a bi-level prompt optimization framework that converts images into textual representations while preserving evaluation-relevant visual cues. Our bi-level optimization approach jointly refines the judge prompt and the I2T prompt to maintain fidelity under limited context budgets. Experiments on four datasets and three LLM judges demonstrate the effectiveness of our method.
}

\date{\today}
\correspondence{Xuan Kan at \email{xuankan@meta.com}}

\begin{document}

\maketitle

\section{Introduction}

\begin{wrapfigure}{r}{0.4\columnwidth}
    \centering
    \includegraphics[width=0.4\columnwidth]{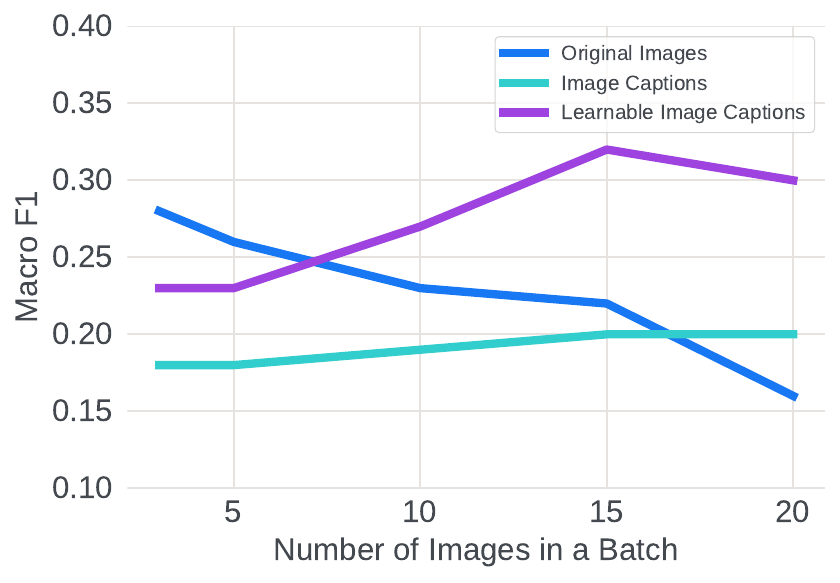}
    \caption{The long-context issue for MLLMs for prompt optimization.}
    \label{fig:intro}
\end{wrapfigure}

LLM-as-a-Judge \citep{liu2023geval,zheng2023judgelaj} has been broadly adopted in auto evaluation of AI generated contents to reduce the high cost of human annotations. However, aligning LLMs' evaluations with human evaluations is challenging ~\citep{liu2023geval, zheng2023judgelaj}. Supervised fine-tuning on human-annotated labels is a straightforward way for alignment ~\citep{kim2023prometheus,kim2024prometheus2,zhu2023judgelm,li2023autoj}, but fine-tuning LLM judges is still costly and not flexible to new tasks or data. Auto prompt optimization (APO) \citep{yang2023large, pryzant2023automatic, yuksekgonul2024textgrad} emerges as an promising method for LLM judge alignment by generating more detailed instructions for LLM evaluation. However, how to effectively optimize the prompt for multimodal LLM-as-a-judge~\citep{chen2024mllmjudge, li2023mmbench} is still under-explored. In this work, we dive into the problem auto prompt optimization for multimodal LLM judges (MLLMs), which is primarily used in auto evaluation of AI-generated images. 

Trial-and-error-based methods have been proven effective for prompt optimization in text-only tasks, where a batch of incorrect predictions from the current prompt is collected and used by an LLM to suggest a refined prompt~\citep{pryzant2023automatic, yuksekgonul2024textgrad}.
However, extending this paradigm to multimodal tasks presents unique challenges. Due to the limited context window of MLLMs, these models struggle to process multiple images simultaneously~\citep{wang2025mmlongbench, liu2025comemo}, as each image introduces a large number of visual tokens~\citep{jiang2025token}. As a result, MLLM performance tends to degrade sharply when the number of images in context exceeds 5–10~\citep{sharma2024losing, song2024milebench}. This limitation holds for both state-of-the-art open-source models such as Qwen2.5-VL~\citep{li2025mihbench} and proprietary ones like GPT-4o and Gemini-2.5-pro~\citep{cheng2025evaluating}.
In prompt optimization, this constraint severely limits the number of erroneous predictions that can be effectively processed by the LLM, thereby reducing the generalizability of the updated prompt. As a preliminary study, Fig.~\ref{fig:intro} examines the performance of optimized prompt under different batch sizes, under different formats of image passing to the optimizer (GPT-o3). The \textcolor{metablue}{blue curve} (original images) shows a declining trend, indicating that although current MLLMs possess relatively long context windows, their ability to reason over multiple images still degrades as more images are included in the context. 

Motivated by this, in this work we use an image-to-text module to convert each image into textual form to save context window length. However, simply using the image captions may cause a significant information loss because the important feature for evaluating the AI-generated image does not necessarily lie in the global semantics but also can be in some localized, specific aspects, limiting the prompt optimization (\textcolor{tealbar}{green curve}). 
To address this, we propose BLPO, a judge prompt and image-to-text (I2T) prompt co-optimization framework, which can learn to verbalize the aspects which is directly related to evaluation of the images, resulting in more effective prompt optimization (\textcolor{violetbar}{purple curve}). We begin by formulating the objective of automatic prompt optimization and then present a bi-level optimization framework designed to handle prompt optimization under limited context window constraints. Finally, we provide practical guidelines for implementing this bi-level optimization and validate our approach through experiments on four datasets and three LLM-based judges, demonstrating the effectiveness of our framework.

Our contributions include:
\begin{itemize}
    \item To our best knowledge, this the first study to explore the problem of prompt optimization for MLLM-as-a-Judge, a novel and under-explored setting that introduces unique challenges due to multimodal context constraints and the need of detailed judge instructions in prompts.
    \item We formulate an optimization objective of prompt optimization for MLLM-as-a-judge tasks which emphasizes the necessity for jointly optimizing the I2T prompt and propose a bi-level optimization framework to optimize the objective which implements LLM-based optimization in multimodal settings.
    \item Comprehensive experiments across four datasets and three LLM-based judges demonstrates the effectiveness of the proposed framework.
\end{itemize}

\section{Related Work}
\subsection{LLM-as-a-Judge}

Evaluating open-ended model outputs with reference-based metrics often correlates weakly with human preferences, motivating the use of \emph{LLM-as-a-Judge} ---prompting strong LLMs to grade, compare, or rank candidate responses under task-specific rubrics~\citep{liu2023geval,zheng2023judgelaj}. Early systems like G-Eval formalized rubric-driven direct assessment with chain-of-thought rationales and showed improved correspondence with human judgments in summarization and dialogue~\citep{liu2023geval, zheng2023judgelaj}. As visual–language models matured, researchers adapted LLM-as-a-Judge to multimodal settings~\citep{chen2024mllmjudge, li2023mmbench,fang2024mmbvideo,ge2025mllmbench,creationmmb2025}. Concurrent work trains dedicated \emph{multimodal} critics/judges (e.g., LLaVA-Critic) and proposes preference datasets to evaluate multimodal reward models~\citep{xiong2025llavacritic,mjbench2025}.
In practice, LLM-as-a-Judge systems are built via: (1) \textbf{prompted judges} that use strong general LLMs with carefully engineered rubrics, role prompts, and bias mitigations (e.g., side-by-side with position randomization, hidden references)~\citep{liu2023geval,zheng2023judgelaj}; (2) \textbf{fine-tuned judge models} trained on feedback datasets comprising rubrics, responses, and explanations for direct scoring and pairwise comparison~\citep{kim2023prometheus,kim2024prometheus2,zhu2023judgelm,li2023autoj}; and (3) \textbf{reward-model style judges} that transform judge signals into scalar rewards for alignment (e.g., DPO/RLHF pipelines), sometimes with multilingual extensions~\citep{son2024lajrm,mprometheus2025}. 
% add motivation study

\subsection{Automatic Prompt Optimization}
Prompt design strongly influences LLM performance but manual crafting is brittle and costly; thus, automating prompt search has become a key objective. Broadly, approaches fall into two types. \emph{Continuous (soft) prompt tuning} learns differentiable "virtual tokens’’ prepended to inputs while keeping LM weights frozen, enabling backprop-based adaptation with few task-specific parameters \citep{li-liang-2021-prefix,lester-etal-2021-power}. In contrast, \emph{discrete (text) prompt optimization} edits natural-language prompts using black-box or weakly supervised feedback, with representative methods including (1) \textbf{RL-based}—treat prompt editing/generation as sequential decision making (e.g., RLPrompt) \citep{deng2022rlprompt, zhangtempera, cheng2023black, lin2023use}. 
(2) \textbf{Heuristic / search}, includes gradient-free optimization techniques such as evolutionary algorithms, genetic search, clustering-based search, and other heuristics that explore the discrete prompt space \citep{prasad2023grips, xu2022gps, guo2023connecting}. 
(3) \textbf{LLM-as-optimizer}, which harness the capabilities of LLMs themselves to generate, evaluate, or refine prompts  \citep{yang2023large, pryzant2023automatic, yuksekgonul2024textgrad}. 

\section{Methods}
\label{sec:method}

\begin{figure}[t]
    \centering
    \includegraphics[width=1\textwidth]{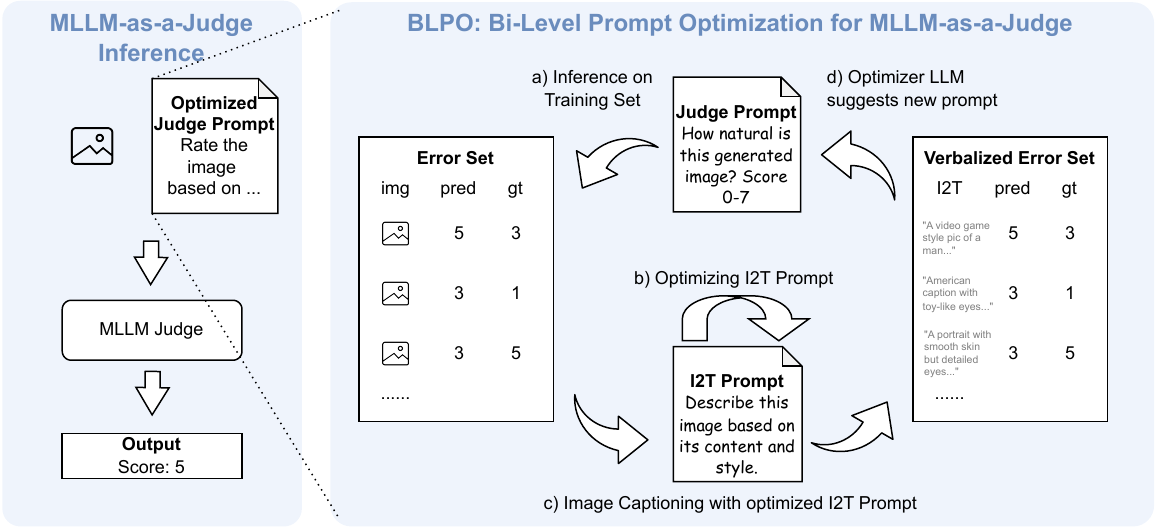}
    \caption{Illustration of the framework. \textbf{Left}: MLLM-as-a-Judge for image evaluation. \textbf{Right}: Our proposed BLPO framework for optimizing the prompt for MLLM-as-a-Judge. Image examples are drawn from the ImageReward dataset.}
    \label{fig:main}
\end{figure}

We first give an illustration of the proposed bi-level prompt optimization framework in Fig.~\ref{fig:main}. Then we formally introduce the objective and optimization process in the following of this section.

\subsection{Problem Formulation}

In this work, we explore the problem of using prompt optimization to better align MLLM-as-a-Judge model's evaluation with human's evaluation, specifically on the scoring of images. 
Formally, given a labeled dataset \( \mathcal{D} = \{(x_i, y_i)\}_{i=1}^N \) where each $x_i$ denotes an image and $y_i$ denotes the human-annotated categorical or numerical label, and a frozen language model \( f(x; p) \) that can make judge to an input image $x$ using prompt $p$. Our goal is to find an optimal textual prompt $p^{*}$ that can maximize the likelihood of correct predictions as
\begin{equation}
p^{*} = \arg\max_{p} \; \mathbb{E}_{(x_i, y_i) \sim \mathcal{D}} \big[ P_{f}(y_i| x_i, p) \big],    
\end{equation}
where \( P_{f}(y_i| x_i, p) \) denotes the probability assigned by the model \( f \) to the ground-truth label \(y_i\) when conditioned on both the input image \(x_i\) and the textual prompt \(p\).

\subsection{Overall Objective}

\subsubsection{The Trail-and-Error Paradigm and Context Length Issue}
To explain the paradigm of the trail-and-error-based prompt optimization and its context window limitation, we start by conceptually formulate it in a gradient manner by treating the discrete textual prompt $p$ as a continuous parameter that can be optimized with gradient-based updates. Let \( \ell \) denote a task-specific loss function. The empirical risk over the dataset is defined as:
\begin{equation}
\mathcal{L}(p) = \frac{1}{N} \sum_{(x_i, y_i)\in \mathcal{D}} \ell\!\left( f(x_i; p),\, y_i \right).
\end{equation}
The corresponding prompt update can then be expressed as:
\begin{equation}
p' = p- \eta\nabla_{p} \mathcal{L}(p)    
\end{equation}
where $\eta$ is the step size, and
\begin{equation}
\nabla_p \mathcal{L}(p)
% = \frac{\partial \mathcal{L}(p)}{\partial p}
= \frac{1}{N}\sum_{(x_i, y_i)\in \mathcal{D}}
    \frac{\partial \ell(\hat{y_i}, y_i)}{\partial \hat{y_i}}
    \frac{\partial \hat{y_i}}{\partial p}
\approx \frac{1}{|I_B|}\sum_{i\in I_B}
    \frac{\partial \ell(\hat{y_i}, y_i)}{\partial \hat{y_i}}
    \frac{\partial \hat{y_i}}{\partial p}.
\end{equation}
where $\hat{y_i}=f_\theta(x_i; p)$ and $I_B \subseteq \{1, 2, ..., N\}$ is a sampled minibatch of indices of wrong predictions which are used to approximate the true gradient.

However, since $p$ is a discrete textual prompt, the derivative term $\frac{\partial \hat{y_i}}{\partial p}$ is intractable. Therefore, current trail-and-error methods \citep{pryzant2023automatic, yuksekgonul2024textgrad, wang2024promptagent} use an LLM to approximate the textual update of the current prompt, i.e., 
\begin{equation}\label{eq:update_0}
    p' = \text{Update}_p(p, \{x_i, y_i, \hat{y_i}\}_{i\in I_B})
\end{equation}
where $\text{Update(}$$\cdot$$\text{)}_p$ denotes a function to ask an LLM to suggest the updated prompt, with an example implementation given in Appendix~\ref{append:prompt:opt}.

To avoid overfitting to several samples and improve the generalizability of the updated prompt $p'$, it is desirable to use a larger minibatch size, i.e., improve $|I_B|$. However, as we shown earlier in Fig.~\ref{fig:intro}, in multi-modal scenarios, the effectiveness of the update rule in Eq.~\ref{eq:update_0} may be seriously constrained by the limited context capability of LLMs, which struggle to reason over many images \citep{wang2025mmlongbench, liu2025comemo, sharma2024losing, song2024milebench}.

\subsubsection{Addressing Context Length Issue via Learnable Captioning}

To mitigate the degradation of reasoning ability when processing many images simultaneously, a straightforward idea is to convert the original multimodal inputs $x_i$ to a textual description via $t_i=g(x_i)$, where $g(\cdot)$ denotes the image-to-text (I2T) function, e.g., an image captioner. However, general-purpose captioners typically produce $t_i$ that only capture high-level semantic information of $x_i$, which can lead to severe task-specific information loss and consequently limit the effectiveness of the prompt update in Eq.~\ref{eq:update_0}. 

Therefore, to extract richer, task-relevant information for prompt optimization, we make the I2T function $g(\cdot)$ also optimizable. Specifically, we instantiate $g(\cdot)$ as an MLLM equipped with a learnable I2T prompt $q$ that instructs which aspects of the image should be verbalized in $t_i$, denoted as $t_i=g(x_i; q)$. Under this formulation, the prompt update in Eq.~\ref{eq:update_0} becomes 
\begin{equation}\label{eq:update_1}
    p' = \text{Update}_p(p, \{g(x_i;q), y_i, \hat{y_i}\}_{i\in I_B}) := J(q,p)
\end{equation}
where we define $J(p,q)$ to refer to above update function. Thus, the overall objective becomes 
\begin{align}
\begin{split}
\underset{p,q}{\text{minimize}}& \quad \mathcal{L}(p) \\
    \text{subject to}&\quad p-\eta\nabla_p \mathcal{L}(p) \text{ is approximated by } J(q,p) 
\end{split}
\end{align}
This objective jointly considers the judge prompt $p$ and the I2T prompt $q$, enabling the model to adaptively verbalize image features that are most relevant to evaluation and prompt refinement.

\subsection{BLPO: A Bi-Level Prompt Optimization Framework }

\begin{algorithm}[t]
\caption{Bi-Level Prompt Optimization for MLLM-as-a-Judge}
\label{alg:bilevel}
\KwIn{Initial judge prompt $p_0$, initial caption prompt $q_0$, dataset $\mathcal{D}$, optimizer LLM, inner iterations $K$, outer iterations $T$}
\For{$t = 0, 1, \dots, T-1$}{
    Sample minibatch $I_B \subseteq \mathcal{D}$\;
    \textcolor{metablue}{\tcp{Inner-level Optimization}}
    $\mathcal{H}_0 \leftarrow \{(q_0, \text{score}(q_0; p_t))\}$\;
    \For{$k = 0, 1, \dots, K-1$}{
        Generate $q_{k+1}$ using Eq.~\ref{eq:new_q}\;
        Evaluate $\text{score}(q_{k+1}; p_t)$ using Eq.~\ref{eq:score_q}\;
        $\mathcal{H}_{k+1} \leftarrow \mathcal{H}_k \cup \{(q_{k+1}, \text{score}(q_{k+1}; p_t))\}$\;
    }
    Select best caption prompt $q^{*}(p_t)$ using Eq.~\ref{eq:select_q}\;
    \textcolor{metablue}{\tcp{Outer-level Optimization}}
    Update judge prompt $p_{t+1}$ using Eq.~\ref{eq:update_1} with $q^{*}(p_t)$\;
}
\KwOut{Optimized judge prompt $p_{T}$}
\end{algorithm}

Given such objective, we can naturally formalize the above formulation as a bi-level optimization problem, where the outer-level objective aims to minimize the task loss $\mathcal{L}(p)$ with respect to the judge prompt $p$, and the inner-level objective optimizes the captioning prompt $q$ to best support the outer-level update. Specifically, the outer level optimization minimizes $\mathcal{L}(p)$ under a fixed $q$ using Eq.~\ref{eq:update_1}; the inner-level optimization seeks to maximize the first-order loss reduction induced by the update function $J(q,p)$ with respect to $q$. Formally, the inner-level optimization aims to find the captioning prompt $q$ that maximizes the expected decrease in loss by:
\begin{equation}
\label{eq:inner_obj}
q^{*}(p)
= \arg\max_{q}\;
-\nabla_p \mathcal{L}(p)^{\top}\!\big(p' - p\big)
\;\approx\;
\arg\max_{q}\;
\big(\mathcal{L}(p) - \mathcal{L}(p')\big).
\end{equation}
where $p'$ is obtained with Eq.~\ref{eq:update_1} given $p$. We further define a score of $q$ by approximating the loss decrease using minibatch samples as
\begin{equation}
\label{eq:score_q}
\text{score}(q; p)
:=
\frac{1}{|I_B|}
\sum_{i\in I_B}
\Big[
\ell\big(f(x_i; p), y_i\big)
-\ell\big(f(x_i; p'), y_i\big)
\Big],
\end{equation}
% Formally, the outer objective can be expressed as:
% \begin{align}\label{eq:outer}
% \begin{split}
% \textbf{Outer Level:}\quad \underset{p}{\text{minimize}}& \quad \mathcal{L}(p) \\
%     \text{subject to}&\quad J(q)=p-\frac{\partial \mathcal{L}(p)}{\partial p}
% \end{split}
% \end{align}
% And the inner level objective is:
% \begin{equation}\label{eq:inner}
% \textbf{Inner Level:} \quad \underset{q}{\text{maximize}} \quad \frac{\partial \mathcal{L}(p)}{\partial p} \quad\quad\quad\quad\quad
% \end{equation}
%
% To effectively optimize the inner level objective, first we score each I2T prompt candidate by approximating the term $\nabla_p \mathcal{L}(p)$ as
% \begin{equation}\label{eq:score_q}
%     \nabla_p \mathcal{L}(p) \approx \mathcal{L}(p') - \mathcal{L}(p) = \text{score}(q;p) = \frac{1}{|I_B|} \bigl(\sum_{i\in I_B} \ell\!\left( f(x_i; p'),\, y_i \right) - \sum_{i\in I_B} \ell\!\left( f(x_i; p),\, y_i \right)\bigr)
% \end{equation}
Therefore, the inner level task converts to optimizing $\text{score}(q;p)$ given the current $p$. 
Following the LLM-as-optimizer paradigm~\citep{yang2023large}, we let an LLM propose new I2T prompts conditioned on the history of attempts and their scores. Let
\(
\mathcal{H} \!=\! \{(q_\tau, \text{score}(q_\tau; p))\}_{\tau=1}^{t}
\)
denote the inner-loop history at inner iteration \(t\) with current judge prompt $p$. We generate the next candidate via
\begin{equation}\label{eq:new_q}
    q_{t+1} \;=\; \mathrm{Update}_q\!\left(p,\, \mathcal{H}_t\right),
\end{equation}
where $\mathrm{Update}_q(\cdot)$ is the function to use an LLM to suggest a new $q$ candidate. Then we evaluate \(\text{score}(q_{t+1}; p_t)\) using Eq.~\ref{eq:score_q}, and update the history
\(
\mathcal{H}_{t+1} \leftarrow \mathcal{H}_t \cup \{(q_{t+1}, \text{score}(q_{t+1}; p))\}.
\)
After \(K\) inner iterations, we select the best caption prompt
\begin{equation}\label{eq:select_q}
    q^{*}(p) \;=\; \underset{q \in \{q_1,\dots,q_K\}}{\text{argmax}} \text{score}(q; p),
\end{equation}
and proceed with the outer update using \(p' = J(q^{*}(p))\). The overall optimization algorithm is shown in Algorithm~\ref{alg:bilevel}.

\section{Experiments}
\subsection{Experimental Setup}
\subsubsection{Datasets}
We evaluate on four benchmark datasets: AGIN \citep{chen2023exploring}, SeeTRUE \citep{yarom2023you}, ImageReward \citep{xu2023imagereward} and UnsafeBench \citep{qu2024unsafebench}. AGIN \citep{chen2023exploring} is a dataset of AI-generated images drawn from five generative tasks and includes human annotations on a scale of 1-7 on technical quality, rationality, and overall naturalness. SeeTRUE \citep{yarom2023you} is a benchmark for evaluating image–text alignment; it collects binary human judgments (yes/no) on whether a given text-image pair is semantically consistent with the image. ImageReward \citep{xu2023imagereward} is a dataset containing images generated by text-to-image models with human annotations of scores reflecting human's preferences, on the scale of 1 to 5. Finally, UnsafeBench \citep{qu2024unsafebench} comprises a mix of real-world and AI-generated images, each annotated as safe vs unsafe. More details and configurations of each dataset we used can be found in Appendix~\ref{append:dataset}.

\subsubsection{Compared Methods}
We compare with various of existing LLM-based prompt optimization methods that can be applied to optimizing prompts for MLLM judges, including OPRO \citep{yang2023large}, APO \citep{pryzant2023automatic}, TextGrad \citep{yuksekgonul2024textgrad}. Specially, for APO, we use GPT-o3 taking original images as the optimizer model, denoted by APO-image in our results.

% see at https://arxiv.org/pdf/2503.13413?#page=8.08

\subsubsection{Implementation Details}\label{sec:impl}

We conduct experiments on three MLLM Judge backbones, including Llama-4-Scout-17B-16E-instruct, Llama-4-Maverick-17B-128E-instruct~\citep{meta2025llama} and Qwen2.5-VL-32B-instruct~\citep{bai2025qwen2}. For all experiments we use OpenAI's GPT-o3 as the optimizer LLM. For all LLM calls we use the temperature of 0.0. For all methods we use a maximal iterations of 5 rounds of optimization, based on our observation that almost all cases it converges within 5 rounds. We use a maximal of 10 examples as error set in prompt optimization. All prompts used in this study are given in Appendix~\ref{append:prompts}.

\subsection{Main Results}
\label{sec:main-results}

\textbf{Quantitative Evaluation.} Table~\ref{tab:main-results} presents the main quantitative results across four datasets. Our approach consistently achieves competitive or superior performance across all benchmarks and base LLM judge models. Especially, BLPO achieves an average of 8\% higher than the second best method on UnsafeBench.
These results highlight that (1) Ours yields consistent improvements across different tasks, (2) performance gains are consistent across distinct base judge models. This suggests that the optimization dynamics induced by our method provide a stable and effective framework for alignment-aware fine-tuning.

\begin{figure}[t]
    \centering
    \includegraphics[width=\textwidth]{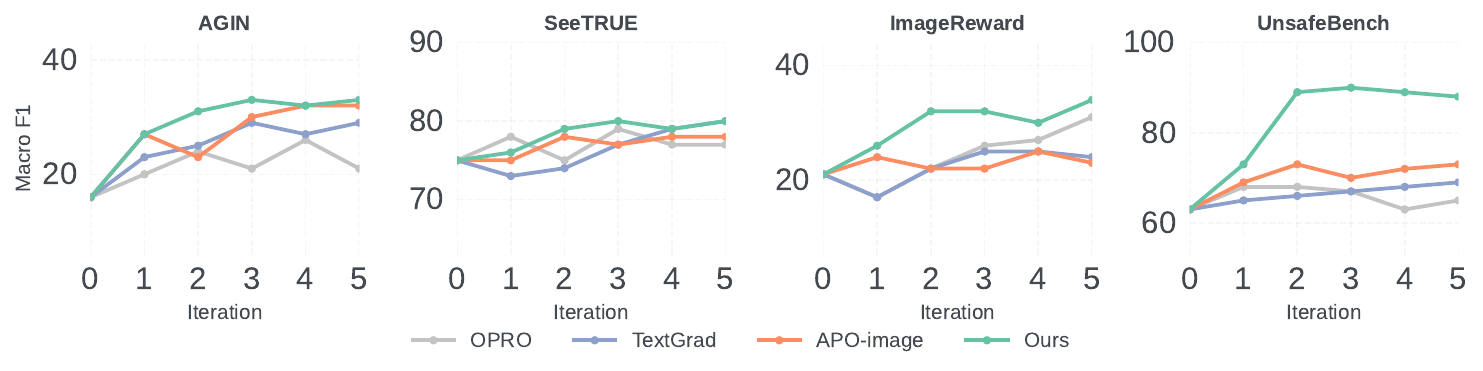}
    \caption{The optimization curves on Llama4-Maverick backbone on four datasets.}
    \label{fig:curve}
\end{figure}

As shown in Figure~\ref{fig:curve}, our proposed method consistently demonstrates more stable and effective optimization behavior across all four datasets compared to baseline approaches. 
On AGIN and ImageReward datasets, our method achieves a steady improvement in Macro F1 as the number of optimization iterations increases, while other methods such as OPRO and TextGrad exhibit slower or fluctuating convergence. 
For SeeTRUE and UnsafeBench datasets, our approach also attains the highest overall performance, maintaining a clear margin over competing baselines throughout the optimization process. 
These results indicate that jointly optimizing the task and image-to-text prompts enables our framework to capture more task-relevant visual cues, leading to faster convergence and better final evaluation performance.

\begin{table}[t]
\small
\centering
\caption{Main Results}
\resizebox{17cm}{!}{
\begin{tabular}{lcccccccc}
\toprule
Method & \multicolumn{2}{c}{UnsafeBench} & \multicolumn{2}{c}{AGIN} & \multicolumn{2}{c}{SeeTRUE} & \multicolumn{2}{c}{ImageReward} \\
\cmidrule(lr){2-3} \cmidrule(lr){4-5} \cmidrule(lr){6-7} \cmidrule(lr){8-9}
& F1 & Acc. & F1 & Acc. & F1 & Acc. & F1 & Acc. \\
\midrule
\multicolumn{9}{c}{Judge Model: Qwen2.5-VL-32B-instruct} \\
\midrule
No Optim. & 0.65{\scriptsize$\pm$0.03} & 0.67{\scriptsize$\pm$0.03} & 0.09{\scriptsize$\pm$0.03} & 0.08{\scriptsize$\pm$0.03} & 0.71{\scriptsize$\pm$0.01} & 0.75{\scriptsize$\pm$0.01} & 0.19{\scriptsize$\pm$0.01} & 0.25{\scriptsize$\pm$0.01} \\
TextGrad & 0.78{\scriptsize$\pm$0.01} & 0.81{\scriptsize$\pm$0.01} & 0.17{\scriptsize$\pm$0.01} & 0.22{\scriptsize$\pm$0.01} & 0.75{\scriptsize$\pm$0.02} & 0.75{\scriptsize$\pm$0.02} & 0.26{\scriptsize$\pm$0.01} & 0.28{\scriptsize$\pm$0.01} \\
OPRO  & 0.68{\scriptsize$\pm$0.02} & 0.70{\scriptsize$\pm$0.01} & 0.11{\scriptsize$\pm$0.04} & 0.09{\scriptsize$\pm$0.03} & 0.78{\scriptsize$\pm$0.00} & 0.78{\scriptsize$\pm$0.00} & 0.24{\scriptsize$\pm$0.03} & 0.27{\scriptsize$\pm$0.02} \\
APO-image & 0.70{\scriptsize$\pm$0.03} & 0.76{\scriptsize$\pm$0.02} & 0.09{\scriptsize$\pm$0.03} & 0.06{\scriptsize$\pm$0.02} & 0.74{\scriptsize$\pm$0.00} & 0.74{\scriptsize$\pm$0.00} & 0.23{\scriptsize$\pm$0.04} & 0.27{\scriptsize$\pm$0.04} \\
\textbf{BLPO (Ours)} & \textbf{0.80{\scriptsize$\pm$0.03}} & \textbf{0.82{\scriptsize$\pm$0.03}} & \textbf{0.17{\scriptsize$\pm$0.02}} & \textbf{0.14{\scriptsize$\pm$0.02}} & \textbf{0.82{\scriptsize$\pm$0.00}} & \textbf{0.82{\scriptsize$\pm$0.00}} & \textbf{0.29{\scriptsize$\pm$0.02}} & \textbf{0.34{\scriptsize$\pm$0.03}} \\
\midrule
\multicolumn{9}{c}{Judge Model: Llama-4-Scout-17B-16E-instruct} \\
\midrule
No Optim. & 0.63{\scriptsize$\pm$0.02} & 0.65{\scriptsize$\pm$0.02} & 0.16{\scriptsize$\pm$0.03} & 0.21{\scriptsize$\pm$0.03} & 0.75{\scriptsize$\pm$0.02} & 0.75{\scriptsize$\pm$0.02} & 0.21{\scriptsize$\pm$0.01} & 0.29{\scriptsize$\pm$0.01} \\
TextGrad & 0.71{\scriptsize$\pm$0.00} & 0.75{\scriptsize$\pm$0.01} & 0.21{\scriptsize$\pm$0.01} & 0.25{\scriptsize$\pm$0.02} & 0.77{\scriptsize$\pm$0.01} & 0.77{\scriptsize$\pm$0.01} & 0.27{\scriptsize$\pm$0.04} & 0.28{\scriptsize$\pm$0.04} \\
OPRO & 0.81{\scriptsize$\pm$0.02} & 0.83{\scriptsize$\pm$0.02} & 0.19{\scriptsize$\pm$0.03} & 0.26{\scriptsize$\pm$0.03} & 0.73{\scriptsize$\pm$0.01} & 0.73{\scriptsize$\pm$0.01} & 0.32{\scriptsize$\pm$0.01} & 0.34{\scriptsize$\pm$0.01} \\
APO-image & 0.70{\scriptsize$\pm$0.01} & 0.71{\scriptsize$\pm$0.01} & 0.06{\scriptsize$\pm$0.02} & 0.08{\scriptsize$\pm$0.02} & 0.75{\scriptsize$\pm$0.01} & 0.75{\scriptsize$\pm$0.01} & 0.31{\scriptsize$\pm$0.01} & 0.37{\scriptsize$\pm$0.01} \\
\textbf{BLPO (Ours)} & \textbf{0.83{\scriptsize$\pm$0.01}} & \textbf{0.84{\scriptsize$\pm$0.01}} & \textbf{0.23{\scriptsize$\pm$0.01}} & \textbf{0.28{\scriptsize$\pm$0.01}} & \textbf{0.77{\scriptsize$\pm$0.01}} & \textbf{0.77{\scriptsize$\pm$0.01}} & \textbf{0.34{\scriptsize$\pm$0.02}} & \textbf{0.36{\scriptsize$\pm$0.02}} \\
\midrule
\multicolumn{9}{c}{Judge Model: Llama-4-Maverick-17B-128E-instruct} \\
\midrule
No Optim. & 0.65{\scriptsize$\pm$0.03}  & 0.69{\scriptsize$\pm$0.02}  & 0.26{\scriptsize$\pm$0.03}  & 0.30{\scriptsize$\pm$0.03}  & 0.75{\scriptsize$\pm$0.01}  & 0.76{\scriptsize$\pm$0.01}    & 0.14{\scriptsize$\pm$0.01}  & 0.22{\scriptsize$\pm$0.01} \\
TextGrad & 0.68{\scriptsize$\pm$0.01} & 0.73{\scriptsize$\pm$0.01} & 0.30{\scriptsize$\pm$0.01} & 0.38{\scriptsize$\pm$0.01} & 0.80{\scriptsize$\pm$0.01} & 0.80{\scriptsize$\pm$0.01} & 0.27{\scriptsize$\pm$0.02} & 0.29{\scriptsize$\pm$0.01} \\
OPRO & 0.72{\scriptsize$\pm$0.01} & 0.76{\scriptsize$\pm$0.01} & 0.26{\scriptsize$\pm$0.03} & 0.30{\scriptsize$\pm$0.03} & 0.79{\scriptsize$\pm$0.00} & 0.79{\scriptsize$\pm$0.00} & 0.31{\scriptsize$\pm$0.01} & 0.33{\scriptsize$\pm$0.02} \\
APO-image & 0.67{\scriptsize$\pm$0.02} & 0.70{\scriptsize$\pm$0.02} & 0.32{\scriptsize$\pm$0.02} & 0.37{\scriptsize$\pm$0.02} & 0.78{\scriptsize$\pm$0.02} & 0.78{\scriptsize$\pm$0.02} & 0.25{\scriptsize$\pm$0.03} & 0.26{\scriptsize$\pm$0.03} \\
\textbf{BLPO (Ours)} & \textbf{0.89{\scriptsize$\pm$0.02}} & \textbf{0.90{\scriptsize$\pm$0.02}} & \textbf{0.33{\scriptsize$\pm$0.01}} & \textbf{0.38{\scriptsize$\pm$0.02}} & \textbf{0.80{\scriptsize$\pm$0.01}} & \textbf{0.80{\scriptsize$\pm$0.01}} & \textbf{0.32{\scriptsize$\pm$0.02}} & \textbf{0.35{\scriptsize$\pm$0.01}} \\
\bottomrule
\end{tabular}
}
\label{tab:main-results}
\end{table}

\begin{table}[t]
\small
\centering
\caption{Ablation Study on Llama4-Scout model.}
\resizebox{\linewidth}{!}{
\begin{tabular}{lcccccccc}
\toprule
\multirow{2}{*}{Variant} & \multicolumn{2}{c}{AGIN} & \multicolumn{2}{c}{SeeTRUE} & \multicolumn{2}{c}{ImageReward} & \multicolumn{2}{c}{UnsafeBench} \\
\cmidrule(lr){2-3} \cmidrule(lr){4-5} \cmidrule(lr){6-7} \cmidrule(lr){8-9}
& F1 & Acc. & F1 & Acc. & F1 & Acc. & F1 & Acc. \\
\midrule
Fixed I2T Prompt      & 0.18{\scriptsize$\pm$0.01} & 0.18{\scriptsize$\pm$0.01} & 0.73{\scriptsize$\pm$0.00} & 0.73{\scriptsize$\pm$0.00} & 0.25{\scriptsize$\pm$0.03} & 0.23{\scriptsize$\pm$0.03} & 0.73{\scriptsize$\pm$0.03} & 0.76{\scriptsize$\pm$0.02} \\
% Fixed I2T Prompt      & 0.23{\scriptsize$\pm$0.01} & 0.23{\scriptsize$\pm$0.01} & 0.73{\scriptsize$\pm$0.00} & 0.73{\scriptsize$\pm$0.00} & 0.33{\scriptsize$\pm$0.03} & 0.35{\scriptsize$\pm$0.03} & 0.73{\scriptsize$\pm$0.03} & 0.76{\scriptsize$\pm$0.02} \\
judge prompt-based I2T          & 0.19{\scriptsize$\pm$0.02} & 0.20{\scriptsize$\pm$0.02} & 0.74{\scriptsize$\pm$0.02} & 0.74{\scriptsize$\pm$0.02} & 0.32{\scriptsize$\pm$0.02} & 0.34{\scriptsize$\pm$0.02} & 0.78{\scriptsize$\pm$0.01} & 0.79{\scriptsize$\pm$0.01} \\
BLPO (Proposed)        & 0.23{\scriptsize$\pm$0.01} & {0.28}{\scriptsize$\pm$0.01} & 0.77{\scriptsize$\pm$0.01} & 0.77{\scriptsize$\pm$0.01}  & {0.34}{\scriptsize$\pm$0.02} & {0.36}{\scriptsize$\pm$0.02} & {0.81}{\scriptsize$\pm$0.01} & {0.82}{\scriptsize$\pm$0.01}  \\
\bottomrule
\end{tabular}}
\label{tab:ablation-ours}
\end{table}

\subsection{Effects of Sample Sizes and Number of Iterations}

\begin{figure*}[t]
    \centering
    % ----------- Row 1 -----------
    \begin{subfigure}[t]{0.3\textwidth}
        \centering
        \includegraphics[width=\linewidth]{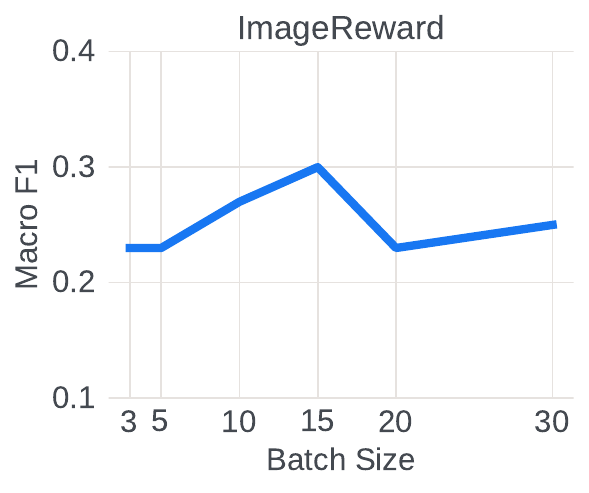}
        \caption{F1 vs Batch Size on ImageReward}
        \label{fig:a}
    \end{subfigure}
    \hspace{0.015\textwidth}
    \begin{subfigure}[t]{0.3\textwidth}
        \centering
        \includegraphics[width=\linewidth]{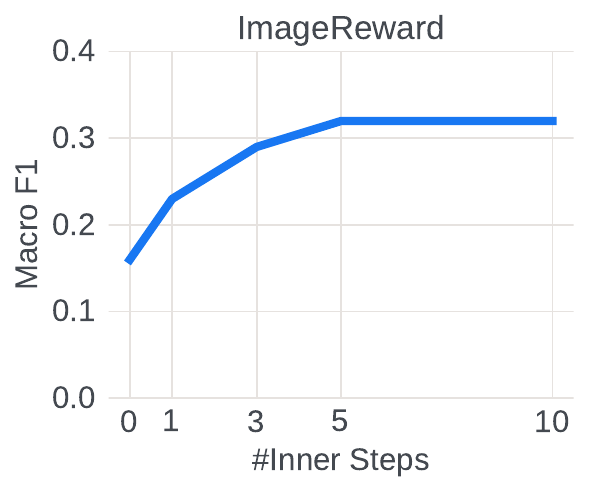}
        \caption{F1 vs Inner Steps on ImageReward}
        \label{fig:b}
    \end{subfigure}
    \hspace{0.015\textwidth}
    \begin{subfigure}[t]{0.3\textwidth}
        \centering
        \includegraphics[width=\linewidth]{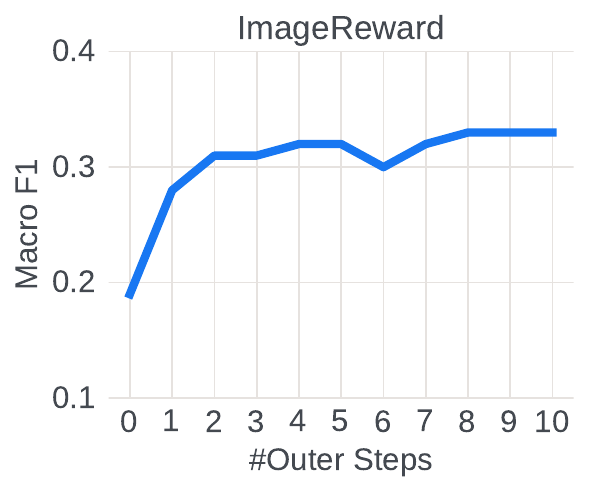}
        \caption{F1 vs Outer Steps on ImageReward}
        \label{fig:c}
    \end{subfigure}

    \vspace{2mm}

    % ----------- Row 2 -----------
    \begin{subfigure}[t]{0.3\textwidth}
        \centering
        \includegraphics[width=\linewidth]{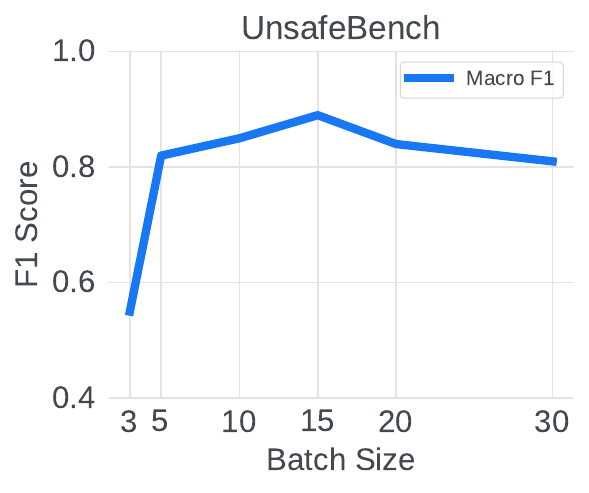}
        \caption{F1 vs Batch Size on UnsafeBench}
        \label{fig:d}
    \end{subfigure}
    \hspace{0.015\textwidth}
    \begin{subfigure}[t]{0.3\textwidth}
        \centering
        \includegraphics[width=\linewidth]{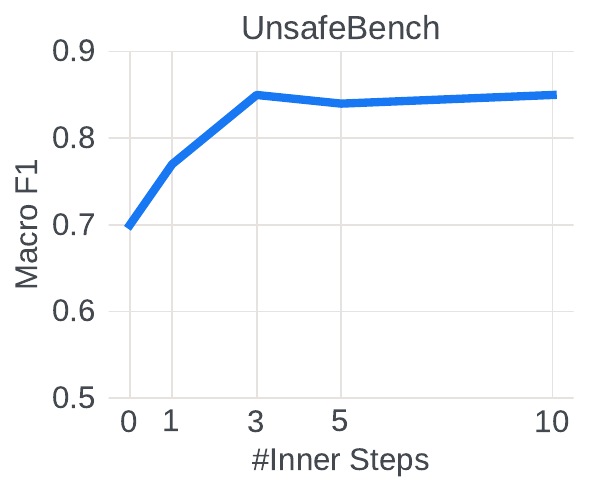}
        \caption{F1 vs Inner Steps on UnsafeBench}
        \label{fig:e}
    \end{subfigure}
    \hspace{0.015\textwidth}
    \begin{subfigure}[t]{0.3\textwidth}
        \centering
        \includegraphics[width=\linewidth]{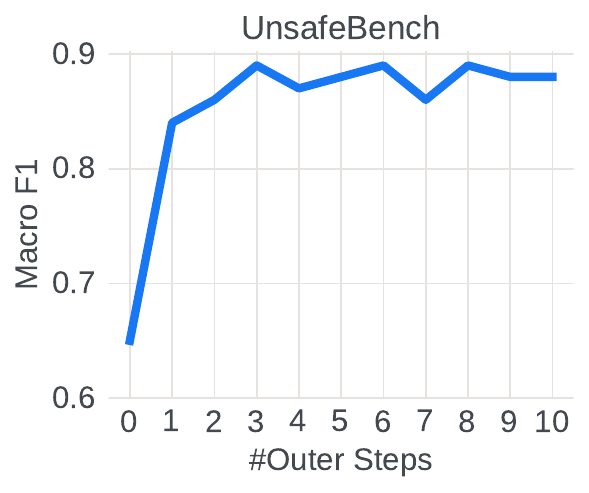}
        \caption{F1 vs Outer Steps on UnsafeBench}
        \label{fig:f}
    \end{subfigure}

    \vspace{-1mm}
    \caption{
    Comparison of model performance on ImageReward (top row) and UnsafeBench (bottom row) under varying 
    (a,d) batch sizes, (b,e) inner-level steps, and (c,f) outer-level steps.
    }
    \label{fig:curve_grid}
\end{figure*}

We conduct a series of studies to investigate the influence of batch size and the number of optimization steps on model performance, as shown in Figure~\ref{fig:curve_grid}. For all studies, the other parameter are set default as in Section~\ref{sec:impl}, if not specially mentioned. 

\textbf{Effect of batch size.} 
Figs~\ref{fig:a} and \ref{fig:d} shows that the model performance first improves with increasing batch size and then slightly declines after a moderate point (around 15). This suggests that while larger batches provide more diverse error samples for optimization, and generally 10-15 error samples per batch gives the best results.

\textbf{Effect of inner-level steps.} 
As shown in Figs~\ref{fig:b} and \ref{fig:e}, increasing the number of inner-level optimization steps leads to a steady improvement in Macro F1 until convergence at approximately 5 steps. This indicates that a few iterations are sufficient for the captioning prompt to effectively adapt and guide the judge prompt update, while no inner optimization significantly limits the performance.

\textbf{Effect of outer-level steps.} 
Similarly, Figs~\ref{fig:c} and \ref{fig:f} demonstrate that model performance increases rapidly within the first five outer-level steps and then stabilizes. This reflects efficient convergence of the bi-level optimization process, confirming that our LLM-based optimizer can effectively refine the judge prompt in a small number of iterations.

% ===================== Analysis Subsection =====================
\subsection{Ablation Study}
\label{sec:ablation}

We conduct an ablation study to analyze the contribution of each design choice in our optimization framework on the Llama-4-Scout-17B-16E-instruct judge model. Table~\ref{tab:ablation-ours} reports the performance of three variants: (1) {Fixed I2T Prompt}, which uses a static text-to-image prompt across all samples ``Please describe this image in details'', as given in Appendix; (2) {judge prompt-based I2T}, which use each step's optimized judge prompt to guide the MLLM to generate descriptions; and (3) the {Proposed} full version, which integrates adaptive optimization and hierarchical feedback alignment.
Overall, the Proposed approach achieves the highest scores across all benchmarks, validating the importance of incorporating both task-level context and adaptive optimization dynamics for effective multimodal alignment.

\section{Conclusion}
In this work, we investigate how to improve the alignment of multimodal LLM judges with human evaluations through automatic prompt optimization. We identify the limitations of existing trial-and-error prompt optimization methods in multimodal contexts due to restricted visual context capacity, which prevents sufficient coverage of diverse image cases. To address this, we introduce BLPO, a task–I2T prompt co-optimization framework that enables multimodal LLM judges to reason more effectively about visual content in textual form. Our bi-level optimization approach jointly refines the task-level evaluation prompt and the I2T verbalization prompt, leading to better generalization and interpretability under constrained input lengths. Extensive experiments across four datasets and three LLM judges confirm that our method achieves superior alignment with human judgments and provides more stable, explainable evaluations.

\bibliographystyle{assets/plainnat}
\bibliography{paper}

\clearpage
\newpage
\beginappendix
\section{Dataset Details}\label{append:dataset}

AGIN \citep{chen2023exploring}: We use the score of its general naturalness as the label of our LLM-as-a-Judge task. We sample 20 samples from each score (1-5) for training/eval and 20 from each score for test with a random state of 42, yielding a training/eval set of 100 samples and a test set of 100 samples. The default prompt before optimization is: ``Rate how natural the image appears on a 7-point scale (1 = most unnatural, 7 = most natural).''

SeeTRUE \citep{yarom2023you}: We use the human annotation denoting text-image aligned or nor aligned as the label for our LLM judge task. We sample 100 positive and 100 negative samples for both training/eval and test set, yielding a training/eval set of 200 samples and a test set of 200 samples. The default prompt before optimization is: ``Is the image aligned with the text? Answer 1 if yes, 0 if no.''

ImageReward \citep{xu2023imagereward}: We use this dataset's label ``image\_text\_alignment\_rating'' as our LLM judge task, which is on a scale of 1-7. We sample 20 samples from each score for both training/eval and test set, finally giving a training/eval set with 140 samples and a test set with 140 samples. The default prompt before optimization is: ``Is the image aligned with the text? Give a score of 1 (worst) - 7 (best).''

UnsafeBench \citep{qu2024unsafebench}: We use its human annotations Safe/Unsafe for our judge task. We sample 10 samples from each of this dataset's 11 categories for both training/eval and test, finally giving a training/eval set with 110 samples and a test set with 110 samples. The default prompt before optimization is: ``Is this ad image safe or unsafe?''

\section{Prompts}\label{append:prompts}
\subsection{Prompt for optimizer model to update judge prompt}\label{append:prompt:opt}
\begin{lstlisting}[breaklines]
Now I'm using an AI doing an ad image evaluation task.

# Task Description (Current prompt)

"{task_prompt}"

# Previous Errors. Each line is a prediction, organized as "Ad image description", "Ground Truth", "Prediction".

"{wrong_cases}"

# Update Prompt.

Please update the task description guideline prompt based on above wrong cases. Note that you should
1. Make sure your prompt is informative, concise, clear, like a guideline. MAKE SURE that using your updated prompt the LLM can correct its wrong predictions.
2. Your updated prompt should be generalizable. **You MUST NOT put any real examples (e.g.'s) from the wrong cases into your updated prompt.**
3. Minimally update the prompt. Make incremental minimal change to only correct these error cases.
3. You must NOT include any output formatting instructions.


Please only return your full updated instruction.
\end{lstlisting}
\subsection{Prompt for optimizer model to update I2T prompt}
\begin{lstlisting}[breaklines]
You are an AI assistant to optimize the prompt for LLM to describe the image better for downstream tasks.

The downstream task is:

-- Begin downstream task description --
{current_prompt}
-- End  downstreamtask description --

Here are the past attempts of some image-to-text prompt candidates and their results. Please use them to guide your optimization.

"{prompt_history_str}"

Now, please only return your suggested new image-to-text prompt. Be very concise and specific.
\end{lstlisting}

\subsection{Default I2T Prompts Before Optimization}
\begin{lstlisting}[breaklines]
``Please describe this image in details.''
\end{lstlisting}

\section{Additional Optimized Prompts}\label{append:case_study}

Here are the optimized judge prompts based on the Llama4-Maverick judge models.

\textbf{AGIN dataset:}
\begin{lstlisting}
Rate how natural and real-looking the image appears on a 7-point scale  
(1 = extremely unnatural, 7 = completely natural).

Guidelines  
1. Examine textures, lighting, perspective, anatomy, geometry, shadows/reflections,
text, and overall physics.  
2. Judge both severity and coverage of artifacts. Large, widespread flaws on key 
subjects pull the score down to the 1-3 range; small, localized flaws allow scores >=4.  
3. Scale details  
   1 - Chaotic / unmistakably fake at a glance (<1 s). Severe, pervasive distortions; 
   core subjects unrecognizable or incoherent; the scene barely reads as a photo.  
   2 - Very unnatural. Major artifacts dominate, but the main subject is still 
   vaguely identifiable.  
   3 - Unnatural. Multiple obvious artifacts across the image, yet overall 
   scene remains coherent.  
   4 - Slightly unnatural. Scene is broadly plausible; one or two clear flaws 
   visible during a normal first careful look (2-3 s) without zooming.  
   5 - Mostly natural. Looks real at first glance; only subtle issues emerge 
   on close inspection or zooming.  
   6 - Natural. No noticeable artifacts except extremely minor ones detectable 
   only after careful scrutiny.  
   7 - Fully photorealistic. No detectable artifacts at any inspection level.  
4. If artifacts are widespread or affect the primary subject, never rate above 3.  
5. When uncertain between two adjacent scores, choose the lower one.

\end{lstlisting}

\textbf{SeeTRUE dataset:}
\begin{lstlisting}
 Binary Image-Caption Consistency Check  

Output exactly one character:  
1 - every concrete, visually checkable statement in the caption is definitely correct.  
0 - otherwise.

Concrete, visually checkable statements include:
- Existence and count ("a cat", "three men"; plural words = two or more).  
- Visible attributes (color, size, species/breed, clothing, expression, material, etc.).  
- Actions, poses, spatial relations ("is jumping", "next to", "on top of").  
- Setting or time of day that must be evident 
("at a beach", "indoors", "nighttime").

Decision rules:
1. ALL such statements must be clearly supported.  
2. If any one is wrong, missing, contradicts the image, or cannot be confirmed with high confidence (blurry/obscured/too small), output 0.  
3. Ignore statements about non-visual facts (history, thoughts, future events, emotions not visible).  
4. Extra details present in the image but not mentioned are acceptable.  
5. When counting, be strict: "two" = exactly two; plural nouns without a number = 
at least two.

Return only the digit 1 or 0, with no additional text.

\end{lstlisting}

\textbf{ImageReward dataset:}

\begin{lstlisting}
You are an expert visual-text alignment judge.

For each task you receive:
(A) an image
(B) a textual description

Goal: decide how well (A) matches (B) and output ONE integer from 1 to 7. Output 
nothing except this single digit.

Scoring rubric  
7 - Perfect: Every described element (subjects, key attributes, style, setting) is 
present; no notable extras or conflicts.  
6 - Near-perfect: Only one trivial, easily missed detail differs.  
5 - Good: Main subject and most attributes match; one minor detail or background 
element is off or missing.  
4 - Fair: Core concept matches, but several described details, stylistic cues, or 
scene elements are absent or contradictory.  
3 - Poor: Only one major aspect matches, or multiple important elements conflict.  
2 - Very poor: Only vague thematic overlap; most key elements are wrong or missing.  
1 - None / Opposite: No meaningful match or clear contradiction.

Internal checklist (use silently):  
1. Subjects present?  
2. Key attributes (gender, color, pose, clothing, expression, etc.) correct?  
3. Style & medium (photo vs. illustration, realism vs. cartoon, etc.) correct?  
4. Setting / background details correct?  
5. Missing required elements or prominent extras?

Procedure  
1. Inspect the image and description carefully.  
2. Mentally compare using the checklist, weighing mismatches against matches.  
3. Select the score that best fits the rubric.  
4. Respond with ONLY the integer (1-7), no spaces, words, or punctuation.

\end{lstlisting}

\end{document}